# Formations Organization in Robotic Swarm Using the Thermal Motion Equivalent Method

Eduard Heiss[1*], Andrey Kozyr[2], Oleg Morozov[3]
[1, 2] Digital Control Systems Laboratory, Tula State University, Tula, Russia
[3] Department of Automatic Control Systems, Tula State University, Tula, Russia
Email: [1] edheiss73@gmail.com, [2] Kozyr_A_V@mail.ru, [3] omo.sau.tsu.tula.ru
*Corresponding Author

*Abstract*— Due to its decentralised, distributed and scalable nature, swarm robotics has great potential for applications ranging from agriculture to environmental monitoring and logistics. Various swarm control methods and algorithms are currently known, such as virtual leader, vector and potential field, and others. Such methods often show good results in specific conditions and tasks. The variety of tasks solved by the swarm requires the development of a universal control algorithm. In this paper, we propose an evolution of a thermal motion equivalent method (TMEM) inspired by the behavioural similarity of thermodynamic interactions between molecules. Previous research has shown the high efficiency of such a method for terrain monitoring tasks. This work addresses the problem of swarm formation of geometric structures, as required for logistics and formation movement tasks. It is shown that the formation of swarm geometric structures using the TMEM is possible with a special nonlinear interaction function of the agents. A piecewise linear interaction function is proposed that allows the formation of a stable group of agents. The results of the paper are validated by numerical modelling of the swarm dynamics. A linear quadrocopter model is considered as an agent. The fairness of the choice of the interaction function is shown.

## I. Introduction

Robotic swarm hold great promise for a variety of applications due to the known qualities of system decentralization, scalability, and resilience to local agent failures [1]–[3]. This paper focuses on the swarm construction of unmanned aerial vehicles (UAVs). A UAV swarm is defined as a cyber-physical system consisting of multiple, possibly heterogeneous UAVs that interact to jointly accomplish a specific mission [4]. The modern uses of swarms are extensive: control [5] and extinguishing [6] fires; fighting insect colonies [7], [8], monitoring agricultural crops [9], monitoring coral reefs [10]; search and rescue operations [11]–[14] environmental monitoring [15]; agriculture [16]; monitoring in the area of mineral extraction [17]; cargo delivery [18], [19]. The extensive application area of autonomous swarms is primarily due to their reliability and the possibility of task distribution. Swarms lead to high efficiency and effectiveness in mission execution, which is impossible to achieve with a single UAV. A large number of papers have been devoted to the study of swarm. The main studies have focused on solving specific theoretical problems such as the organization of a sustainable swarm [20].

A large amount of work is devoted to group control methods. One of the most common methods for a swarm control is the artificial potential field model [21]–[24]. The artificial potential field method [25], [26] provides the construction of a motion trajectory for each element of the formation. This approach organizes the agents` movement and prevents their physical collision. The method is based on the creation of a virtual potential field that attracts the agent to the target and repels it from obstacles. The method has disadvantages, one of them is the occurrence of local minima [27], [28]. However, there are methods to solve such problems, for example, by considering only the nearest objects [29] or by specialised algorithms [30]. A vector field is also used for group control. A vector field is a space, each point of which corresponds to a vector with a given length, which allows to ensure the movement of agents along the required trajectory [31], [32].

A consensus approach is used for control of a large-scale swarm [33]–[35]. In this case, according to a certain control law, all swarm agents converge to the same state [36], [37].

A relevant area of research on swarm is behavioral algorithms that model biological behavior [38], [39]: the particle swarm method [40], which can be used in conjunction with other methods [41]; algorithms inspired by the organization of fish movement [42]; algorithms using virtual pheromones [43]; artificial bee swarm algorithms [44]; and modeling of natural systems, such as the thermal motion equivalent method (TMEM) [45]–[47]. The idea of the TMEM is the organization of the swarm agents` behavioral similarity to the atoms` thermal motion. In this approach, swarm behavior can be controlled by integral parameters equivalent to similar criteria of a thermodynamic system, such as enthalpy, temperature, pressure, etc. The concept of thermal motion of atoms allows UAVs with different initial states to calculate the expected swarm trajectory and avoid collisions and obstacles during formation flight.

Often heuristic algorithms are used to construct trajectories for agents to move. Spatial optimisation of trajectories [48], [49] and spatial-temporal optimisation [50] are also used to construct trajectories for swarm agents. Trajectory construction has received a lot of attention [51]–[55], including disturbance flight issues [56].

A decentralised behaviour-based formation control algorithm is presented in [57]. The paper assumes that all robots have information about the final destination, the geometric structure, the relative position of neighbouring robots and obstacles. The anticipation angle is used to avoid collisions. The disadvantage is the low stability of the formation.

One of the important problems in the swarm control is the formation. There is a simple to implement and widely used "leader agent" approach [58], [59]. In [60], an example of a mathematical model of formation motion in curvilinear coordinates relative to the leader is given. In this method, there is no informational feedback from the agents` formation. The failure or disappearance of the leader irreversibly affects the whole swarm.

The virtual leader application compensates this disadvantage. In [61], the control algorithm is based on the approach with feedback between agents and the formation of a virtual leader in the geometric center of the structure. The UAVs exchange (can exchange) information with each other and with the virtual leader, which is embedded in the network topology subsystem. This approach provides fast dynamic response and low tracking error. However, each UAV in this method is treated as a particle and its shape is not considered.

Many current studies focus on the issues of practical implementation of a swarm UAV system. Methods of agent communication in a swarm are considered in [62], machine learning-based approaches for optimizing agent communication and new routing protocols with dynamic topology properties are proposed. The problems of swarm system navigation are discussed in [63].

The presented brief review of the current state-of-the-art of UAV swarms shows that there is currently a great interest in such systems. Obviously, the application of swarms in various fields is promising. However, there are currently no engineered swarms that solve practical problems and provide benefits. This is largely due to the fact that there is currently no universal approach to UAV swarm formation and control. Most current works consider one or a few tasks and components of the swarm, neglecting the others [64], [65].

In some cases, for example when combining heterogeneous agents into structures, the swarm will require the formation of geometric structures. Unlike others, the thermal motion equivalent method at this stage does not have components that realise formations. Therefore, this paper focuses on improving the TMEM to provide formations.

## II. PROBLEM STATEMENT

The forming geometric structures problem (formations) is reduced to maintaining the predetermined by the operator distance and direction of motion to a neighboring agent. Two agents whose control is aimed at keeping a required distance between them, are called coupled agents. In this paper, the control object is the swarm agent, a quadrocopter. Fig. 1 shows a formation of three agents.

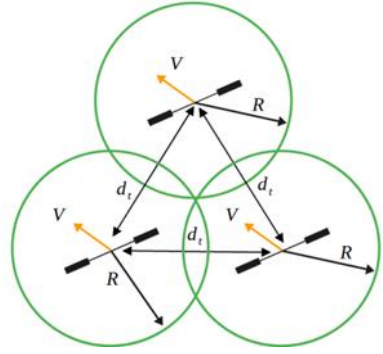

Fig. 1. A formation of three agents

In Fig. 1, $R$ is the UAV interaction radius, $d_t$ is the required distance between UAVs, $V$ is the swarm agent velocity.

In order to evaluate the possibility of coupling agents functioning by TMEM, the quadrocopter motion is considered along the horizontal axis using its simplified model:

$$\begin{cases} \omega_d = k_p(\varphi_d - \varphi); \\ \dot{\omega} = k_d(\omega_d - \omega); \\ \ddot{x} = \varphi \cdot g, \end{cases} \quad (1)$$

where, $\varphi$ is a tilt quadrocopter angle from the vertical (rad); $\varphi_d$ is a required tilt quadrocopter angle from the vertical (rad); $\ddot{x}$ is an acceleration along the x-axis (m/s2).

The quadrocopter linear model (1) is obtained under the conditions:

- zero yaw speed;
- dynamics of propeller groups is not taken into account due to their significant rapidity relative to other inertial elements of the control object;
- the tilt quadrocopter angle from the vertical is small, which allows omitting trigonometric nonlinearities of kinematic relations.

The dynamics of rotational and translational motion along the horizon is described in state space as:

$$\begin{aligned} \dot{x} &= Ax + BU, \\ y &= Cx + DU, \end{aligned} \quad (2)$$

where

$$A = \begin{pmatrix} 0 & 1 & 0 & 0 \\ 0 & 0 & g & 0 \\ 0 & 0 & 0 & 1 \\ 0 & 0 & -k_p k_d & -k_d \end{pmatrix}, B = \begin{pmatrix} 0 \\ 0 \\ 0 \\ k_p k_d \end{pmatrix},$$

$$C = (0 \quad 1 \quad 0 \quad 0), D = (0).$$

### A. Synthesis of the control algorithm

A methodology for the synthesis of a modal controller that provides stabilisation of the RMS velocity of the interacting agents before and after interaction is proposed in [26]. The control is performed as a feedback on all state variables with a vector of gain coefficients $K$:

$$U = Kx. \quad (3)$$

There are considering the quadrocopter motion only along the horizontal axis, so the feedback coefficients are defined as:

$$K(p, g, k_p, k_d) = \begin{pmatrix} -\beta g^{-1}(p_1 p_2 p_3 + p_1 p_2 p_4 + p_1 p_3 p_4 + p_2 p_3 p_4) \\ \beta g^{-1} p_1 p_2 p_3 p_4 \\ \beta(p_2 p_3 + p_2 p_4 + p_3 p_4 + p_1(p_2 + p_3 + p_4)) \\ -\beta(k_d + p_1 + p_2 + p_3 + p_4) \end{pmatrix}, \quad (4)$$

$$\beta = \frac{1}{k_p k_d}.$$

where $p$ are the characteristic equation roots of the open loop system and defined as:

$$p = \begin{pmatrix} -r_l - im_l \cdot i \\ -r_l + im_l \cdot i \\ -im_r \cdot i \\ im_r \cdot i \end{pmatrix}, \quad (5)$$

where $r_l$ is the real part of the left pair of conjugate roots of the characteristic equation; $im_l$ is the imaginary part of the left pair of conjugate roots; $im_r$ is the imaginary part of the right conjugate roots.

*B. Agents interaction*

The difference between the coordinates of the neighbouring quadrocopter $j$ and its own coordinates, corrected for the feedback of the modal controller, is used to compute agent interaction:

$$d = P_j^* - P^*,$$
$$P^* = (P \quad V \quad \varphi \quad \dot{\varphi}) \left(\frac{K}{K_1}\right). \quad (6)$$

The interaction intensity with a neighboring agent j, which is a measure of the sphere's intersection, is defined as:

$$c_j = d - \text{sign}(d)(R + R_j). \quad (7)$$

Then the classical repulsion function, whose value is used as $U$ in (3), depends on the measure of intersection of the spheres, is shown in Fig. 2 and is defined as:

$$C_{des\,j} = \begin{cases} \max(\text{sign}(d_j) C_{\max}, K_1 c_j), \\ \quad if \; |d_j| < R + R_j; \\ 0, otherwise, \end{cases} \quad (8)$$

where $C_{\max}$ is the maximum control value. A type of piecewise linear repulsion function in the TMEM is shown in Fig. 2.

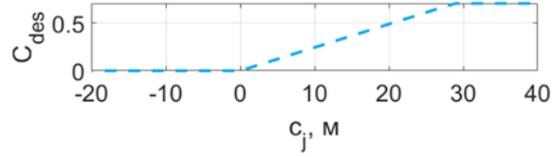

Fig. 2. Push-out functions in the TMEM

The repulsion function shown in Fig. 2 avoids the collision of swarm agents. The effectiveness of using such a function was demonstrated in [26]. However, a modification of this function is required to organize the swarm formation.

### III. SYNTHESIS OF INTERACTION FUNCTION FOR ORGANIZATION OF SWARM SYSTEM FORMATION BY TMEM

*A. Interaction function with attraction*

Equations (1), (5)-(9) describe a model of an agent functioning by TMEM. The TMEM is based on the potential field method. For the forming a geometrical structure and keeping a given distance between agents, the interaction function is refined with "attraction" (Fig. 3).

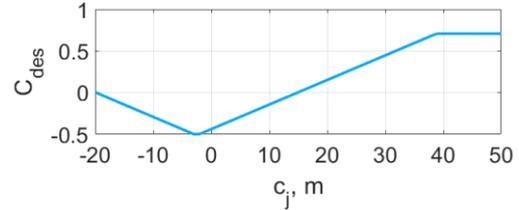

Fig. 3. Interaction function with attraction

Fig. 3 shows that the interaction function of the swarm agents takes a negative value. This condition will encourage the agents to attract each other. Let's consider the effectiveness of the proposed interaction function using a model example.

*B. Swarm simulation*

Numerical simulations of the linear model, whose parameters are given in Table 1 and initial conditions in Table 2, were carried out to demonstrate the performance of the interaction function shown in Fig. 3

TABLE I. PARAMETERS OF MODELING

| Parameter | Value | Parameter | Value |
| --- | --- | --- | --- |
| $k_p$ | 6.0 | $r_l$ | 12.0 |
| $k_d$ | 25.0 | $im_l$ | 0.1 |
| $g$ | 9.8 | $im_r$ | 0.55 |
| $R, R_j$ | 20.0 | - | - |

TABLE II. INITIAL CONDITIONS

| Parameter | Description | Value |
| --- | --- | --- |
| $P_1^0$ | Coordinate of the first agent, m | 50.0 |
| $P_2^0$ | Coordinate of the second agent, m | 0.0 |
| $V_1^0$ | Speed of the first agent, m/s | -1.5 |
| $V_2^0$ | Speed of the second agent, m/s | 3.0 |

A simulation dynamic model of the motion of several quadcopter-type UAVs was developed to investigate the swarm formation with a TMEM. The principle scheme of UAV motion is shown in Fig. 4.

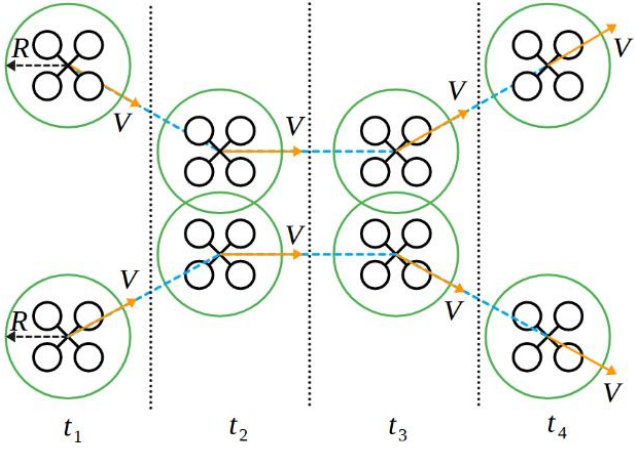

Fig. 4. Dynamic model of swarm formation by the TMEM

The interaction function (Fig. 3) generates excessive maneuvering of the agent - acceleration to the neighboring agent during the interaction. The speed of two quadcopter-type UAVs and their RMS speed are shown in Fig. 5.

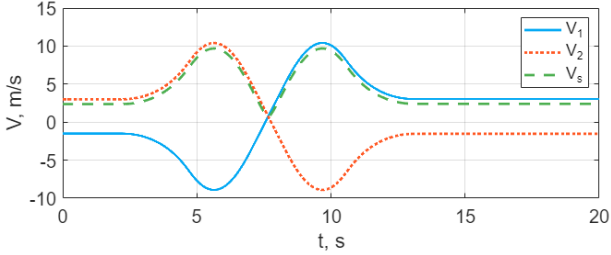

Fig. 5. Result of modeling the interaction between two agents using the attraction function (8)

The results show that there is no coupling with this interaction function.

*C. Interaction function with switching*

A local modal regulator is integrated into the agent for modal functioning. Such a regulator keeps the RMS velocity constant before and after interaction of swarm agents. It is proposed to switch the interaction function without "attraction" to a modified "attraction" function when the agents reach a given distance to ensure pairing and maintain RMS velocity. The modified interaction function is shown in Fig. 6.

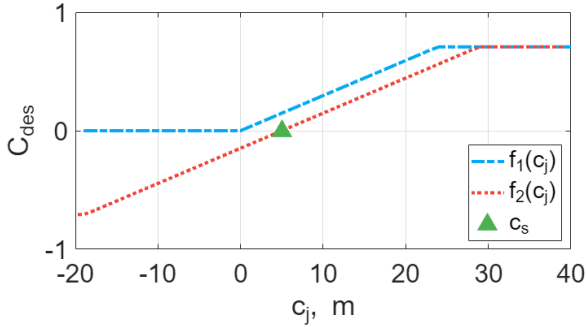

Fig. 6. Modified cooperation function of swarm agents

In Fig.6, $f_1(c_j)$ is the cooperation function without pairing, $f_2(c_j)$ is the cooperation function during pairing, $c_s$ is the switching point between $f_1(c_j)$ and $f_2(c_j)$.

The following is a description of the agents` interaction with the modified function.

In the first moments of agents` interaction when the spheres cross, their acceleration is determined by the function $f_1(c_j)$, $c_j$- the measure of the intersection of spheres (8).

When the agents reach the $c_s$ point, the interaction function is switched from $f_1(c_j)$ to $f_2(c_j)$. The indicator for the switching state is $F_{en}$. Further agents` interaction takes place according to the function $f_2(c_j)$.

The reverse switching from function $f_2(c_j)$ to $f_1(c_j)$, i.e., the uncoupling of agents, is determined by external conditions, e.g., by an operator's command. The uncoupling must take place at the point when:

$$\begin{aligned} \bigl||d| - d_t\bigr| < \varepsilon, \\ d_t < R + R_j, \end{aligned} \quad (9)$$

where $d_t$ is the required distance of coupling between the agents, m; $\varepsilon$ is the switching point neighbourhood width, m. The agent has a certain kinetic energy and energy equivalent to the value of the repulsion function at coupling, so arbitrary switching is inadmissible. Uncoupling must occur at the moment when the kinetic energy of the agent and the difference between the values of functions $f_1(c_j)$ and $f_2(c_j)$ will be equal to those that were at the moment of coupling.

Control in case of "attraction" by means of switching functions has the form

$$\begin{aligned} C_{des\ j} &= \begin{cases} f_1(c_j), & \text{if } F_{en} = 0; \\ f_2(c_j) & \text{otherwise}, \end{cases} \\ f_1(c_j) &= \begin{cases} \max(s_d C_{max}, K_1 c_j), & \text{if } |d_j| < R + R_j \\ 0 & \text{otherwise}, \end{cases} \\ f_2(c_j) &= \max\bigl(s_d C_{max}, K_1(d_j - s_d d_t)\bigr), \\ d_j &= c_j + s_d(R + R_j), \\ s_d &= \text{sign}(d_j). \end{aligned} \quad (10)$$

The numerical modelling parameters and initial conditions are given in Table 1 and Table 2. The uncoupling command occurs at time $t = 29\ s$. The results of the simulation are presented in Fig. 7.

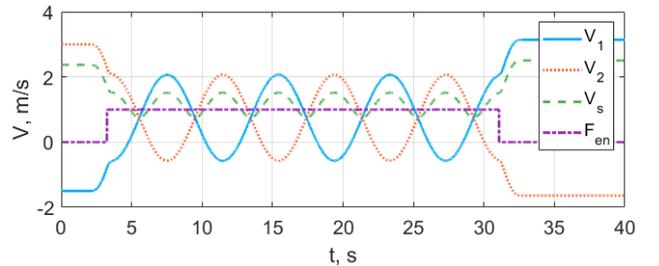

Fig. 7. Result of modeling two agents using the interaction function with switching

The graph shows that both coupling and uncoupling take place. The results show that the actual disconnection occurs at time $t > 30\,s$, while the command occurs at time $t = 29\,s$, i.e. the disconnection is correct. There are agents` oscillations relative to each other at the average distance $d_t$ during coupling at $F_{en} = 1$. The oscillations presence is due to the peculiarity of the modal regulator implementation and the synthesis problem formulation – the RMS agents` velocity before and after the interaction must remain constant.

The change in the RMS agents` velocity in this example before and after the interaction is approximately 5.6%. The results are similar under other initial conditions. This comes from the fact that the interaction function is piecewise linear and has a discontinuity at the coupling point. The control changes stepwise when switching $f_1(c_j)$ and $f_2(c_j)$ back and forth, which is a deviation from the linear model and was not considered in the synthesis of the modal controller.

### D. Improved interaction function with switching

The paper proposes to change the interaction function according to Fig. 8 to reduce the influence of switching.

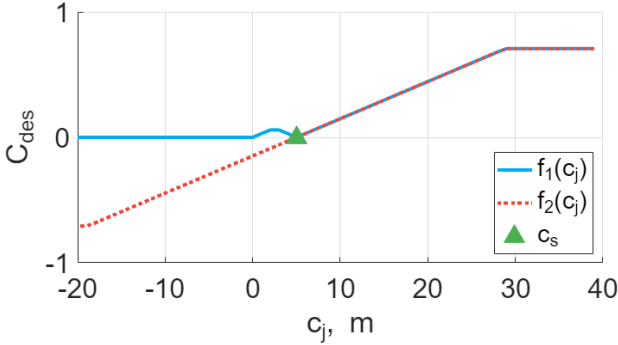

Fig. 8. Improved interaction function with switching

For the variant of the interaction function illustrated in Fig. 6, $f_1(c_j)$ and $f_2(c_j)$ at the coupling point $c_s$ and to the right coincide. Therefore, switching between $f_1(c_j)$ and $f_2(c_j)$ does not cause a step change in control. The control in this case is defined as:

$$C_{des\,j} = \begin{cases} \max\left(s_d C_{max}, f_1(c_j)\right), & if\ F_{en} = 0; \\ f_2(c_j) & otherwise, \end{cases}$$

$$f_1(c_j) = \begin{cases} 0, if\ |d_j| \geq R + R_j; \\ K_1 c_j, if\ b < |d_j| < R + R_j; \\ -K_1(d_j - s_d d_t), if\ d_t \leq |d_j| < b; \\ K_1(d_j - s_d d_t), if\ d_t > |d_j|, \end{cases} \quad (11)$$

$$f_2(c_j) = \max\left(s_d C_{max}, K_1(d_j - s_d d_t)\right),$$

$$b = \frac{d_t + R + R_j}{2},$$

$$d_j = c_j + s_d(R + R_j),$$

$$s_d = \text{sign}(d_j).$$

The numerical modelling parameters and initial conditions are given in Table 1 and Table 2. The uncoupling command also occurs at time $t = 29\,s$. The results of simulation are presented in Fig. 8.

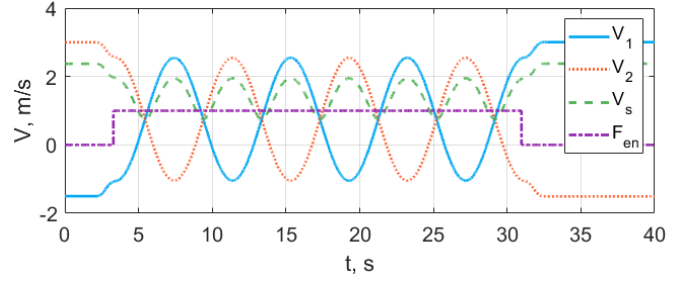

Fig. 9. Result of modeling two agents using the improved interaction function with switching

The amplitude of the oscillations during coupling has increased compared to the previous realisation of the interaction function (10). This is due to the smaller values of the function to the left of the coupling point $c_s$ in Fig. 8. The actual uncoupling also occurs at time $t > 30\,s$, i.e., the uncoupling is correct. It should be noted that it is acceptable to switch from $f_2(c_j)$ to $f_1(c_j)$ not only in the vicinity of the coupling point $c_s$, but also to the right of it (Fig. 8), since to the right of $c_s$ the functions $f_1(c_j)$ and $f_2(c_j)$ coincide in the case of using (11)

The change in the RMS velocity of the agents in this example before and after the interaction is about 0.2%, which is an improvement over to the previous implementation of the interaction function (11).

## IV. CONCLUSION

The paper shows the possibility of organising a UAV swarm formation. We propose an interaction function that ensures the coupling of agents at a predetermined distance with a minimum change in RMS velocity before and after the interaction, consistent with the thermal motion concept. The paper also provides recommendations for changing the repulsion function to a function that causes the agents to oscillate relative to each other at a predetermined distance.

The performance quality of the interaction function and switching between its parts was evaluated by numerical simulation results using a linear quadrocopter model.

Further work is aimed at synchronising the speed of the coupled agents to eliminate oscillations while maintaining RMS speed.

### ACKNOWLEDGMENT

The study was supported by a grant from the Russian Science Foundation No. 23-29-10077, https://rscf.ru/project/23-29-10077/».